\documentclass{article}

\newcommand{\bfml}{}

\usepackage[utf8]{inputenc} 
\usepackage[T1]{fontenc}    
\usepackage{hyperref}       
\usepackage{url}            
\usepackage{booktabs}       
\usepackage{amsfonts}       
\usepackage{nicefrac}       
\usepackage{microtype}      
\usepackage{xcolor}         
\usepackage{amsmath}
\usepackage{amssymb}
\usepackage{graphicx}
\usepackage{algorithm}
\usepackage{algpseudocode}
\usepackage{listings}
\usepackage{multirow}

\newcommand{\BibTeX}{\rm B\kern-.05em{\sc i\kern-.025em b}\kern-.08em\TeX}
\usepackage{amsthm}
\usepackage{bbm}
\usepackage{placeins}
\usepackage{caption}

\newtheorem{thm}{Theorem}

\newtheorem{definition}[thm]{Definition}

\DeclareFontFamily{U}{fsy}{}
\DeclareFontShape{U}{fsy}{m}{n}{<->s*[.9]psyr}{}
\DeclareSymbolFont{der@m}{U}{fsy}{m}{n}
\DeclareMathSymbol{\der}{\mathord}{der@m}{182}

\DeclareSymbolFont{der@m}{U}{fsy}{m}{n}
\DeclareMathSymbol{\derdelta}{\mathord}{der@m}{100}

\DeclareFontFamily{OMS}{smallo}{}
\DeclareFontShape{OMS}{smallo}{m}{n}{<->s*[.65]cmsy10}{}
\DeclareSymbolFont{smallo@m}{OMS}{smallo}{m}{n}
\DeclareMathSymbol{\smallo}{\mathord}{smallo@m}{79}

\DeclareSymbolFont{imag@m}{OT1}{cmr}{m}{ui}
\DeclareMathSymbol{\imag}{\mathord}{imag@m}{105}

\DeclareFontFamily{U}{mathb}{\hyphenchar\font45}
\DeclareFontShape{U}{mathb}{m}{n}{
      <5> <6> <7> <8> <9> <10> gen * mathb
      <10.95> matha10 <12> <14.4> <17.28> <20.74> <24.88> matha12
      }{}
\DeclareSymbolFont{mathb}{U}{mathb}{m}{n}
\DeclareMathSymbol{\monus}{2}{mathb}{"01}

\title{{\tt humancompatible.interconnect}: Testing Properties of  Repeated Uses of Interconnections of AI Systems}

\author{%
  Rodion Nazarov \\
  \footnotesize
  Czech Technical University in Prague \\[6pt]
  \bfml
  Anthony Quinn, Robert Noel Shorten \\
  \footnotesize
  Imperial College London \\[6pt]
  \bfml
  Jakub Mareček \\
  \footnotesize
  Czech Technical University in Prague \\[6pt]
}

\begin{document}

\maketitle


\begin{abstract}
Artificial intelligence (AI) systems  often interact with multiple agents. The regulation of such AI systems often requires that {\em a priori\/} guarantees of fairness and robustness be satisfied.  With stochastic models of agents' responses to the outputs of AI systems, such {\em a priori\/} guarantees require non-trivial reasoning about the corresponding stochastic systems. 
Here, we present an open-source PyTorch-based toolkit for the use of stochastic control techniques in modelling interconnections of  AI systems and properties of their repeated uses. It models robustness and fairness {\em desiderata\/} in a closed-loop fashion, and provides {\em a priori\/} guarantees for these interconnections. The PyTorch-based toolkit removes much of the complexity associated with the provision of fairness guarantees for closed-loop models of multi-agent systems.
\end{abstract}

\section{Introduction}

Our human experience is increasingly influenced by artificial intelligence (AI) systems. 
Regulation is now calling for \emph{a priori} guarantees of fairness and robustness \cite{AIACT} to be met during AI-human interactions. Some pre-determined measure of robustness must be achieved--- with some appropriate degree of statistical certainty ---when designing an AI system, or an interconnection of AI systems. In this approach, the agent will pre-specify the desired fairness or robustness measure and other objectives (accuracy, runtime), along with the desired level of statistical certainty required in achieving them. They will also pre-specify the range of machine learning models and parameter settings. Technical advances in stochastic optimization and control enable delivery of these {\em a priori\/} guarantees {\em in tandem\/} with robust performance of the machine learning training algorithm itself. 

It has recently been pointed out \cite{10555099} that the {\em desiderata\/} of equal treatment and equal impact---as codified in the Civil Rights Act of 1964 \cite{CivilRightsAct}, and in much of the subsequent civil rights legislation \cite{CivilRightsMore}---could be defined using closed-loop models. One pass around the loop makes it possible to define equal treatment, while  long-run-average behaviour around the loop is associated with equal impact. Therefore, the long-run average needs to exist, a requirement that can be defined in terms of the ergodic properties of the closed-loop AI-agent system
 \cite{10555099}. Necessary and sufficient conditions for these ergodic properties to be satisfied had not been known until recently \cite{Hairer2011}. When these ergodic properties are satisfied in a stochastic model of the closed-loop system, one can then reason about fairness of the long-run averages.   

Proofs of robustness and fairness of stochastic models of multi-agent systems are difficult to establish for reasons of complexity. Indeed,  state-space approaches to supervision and verification of modular discrete-event systems are PSPACE-complete \cite{papadimitriou1987complexity,rohloff2005pspace} even in deterministic calculi, and are undecidable \cite{papadimitriou1987complexity,madani2003undecidability} in some stochastic calculi. 
Nevertheless, Hairer's Fields-medal-winning results \cite{Hairer2011} can be interpreted as sufficient (and in some cases necessary) conditions for these ergodic properties to be satisfied.  
Subsequently, Marecek, Shorten {\em et al.}\ \cite{Fioravanti2019,marecek2021predictability,kungurtsev2023ergodic} developed guarantees for stochastic closed-loop models of multi-agent systems, utilizing non-trivial conditions from non-linear control theory (incremental input-to-state stability), along with the results of Hairer and earlier results from applied probability (contractivity on average \cite{Ramen2019}). These allow for the study of the ergodic properties of the multi-agent systems, and, consequently, fairness properties from the individual point of view.

The challenge of proving \emph{a priori} guarantees of fairness and robustness goes  beyond computational complexity, however. 
Often, the interconnections of AI systems can be challenging to model. 
Consider, for example, sharing-economy platforms such as {\em Uber\/} and {\em Airbnb}, and virtual power plants, such as {\em Tesla Virtual Power Plant\/} in the US and {\em Next Kraftwerke\/} in Europe.
In the past decade, many such platforms have been deployed at scale.  Notably, the AI systems in such platforms not only provide decision support, but actually execute actions --- i.e.\ ``perform actuation'' \cite{alleyne2023control} --- with very little human oversight, due to the conflicting latency requirements of the platforms and the human oversight. 
At the same time,  there is great interest in regulating such platforms (for example, renewable energy communities and citizen energy communities in the Renewable Energy Directive \cite{RenewableDirective} and in the Internal Electricity Market Directive of the EU, as amended in \cite{EnergyMartket}), which necessitates multi-agent system modelling. 

The modelling challenge also involves multiple systems with multiple developers and deployers. Consider, for example, AI systems in transportation engineering:
\begin{itemize}
\item vehicle routing, driving assistance, and automated driving systems in privately owned cars;
\item matching algorithms for ride-hailing and car sharing platforms;
\item traffic signals and speed limits being set by AI systems run by transport authorities.  
\end{itemize}
These AI systems are becoming increasingly \emph{interconnected}.
For example, vehicle routing and driving assistance systems can rely on information from traffic control signal (TCS) systems. 
Similarly, speed limits can be adjusted based on the observed speeds of vehicles equipped with automated-driving features. 
These interconnections can improve the efficiency of the system, but they can also affect the robustness and overall performance of the transportation system.
For example, vehicle routing based on traffic control information could help avoid long, repeated stops at intersections. 
We would like to certify that delays---or their remediation---are  fairly distributed among the identified groups in the population.  
Notice, however, that the fairness depends on the existence of long-run averages of the delays, which is not automatic. 

In this article, we present a toolkit---{\tt interconnect}---which enables reasoning about the interconnections of such AI systems implemented on top of PyTorch. The guarantees of ergodic properties are obtained using automated differentiation within PyTorch (autograd), making it possible to reason about a variety of fairness notions.
This removes much of the complexity associated with the provision of fairness guarantees for closed-loop models of multi-agent systems. {\tt interconnect} is made available as an open-source toolkit,  accessible via {\tt humancompatible.org}. The latter is the website of {\em AutoFair}, the Horizon Europe research consortium  on human-compatible AI with fairness guarantees.

\section{A Closed-Loop View of AI Systems}
\label{sec:setup}

We revisit the closed-loop model of \cite{10555099}, and extend it to interconnections of AI systems. Throughout, we utilize the following constraints:
\begin{itemize}
\item The agents do not communicate with one another, or only do so in response to information broadcast by the AI system.
\item Agents obtain information from the AI system, but are not required to take action based on the outputs of the AI system. It will be convenient to encode agents' reactions to the output of the AI system probabilistically, as one Markov state process per agent. 
\item The AI system does not necessarily monitor individual agents' actions (``profiling'' in the language of \cite{AIACT}), but rather some aggregate or otherwise filtered version. 
\item Conditional stochastic independence is adopted as a sufficient condition to ensure fairness~\cite{Lang:24} in the way the AI system interacts with the agents. Specifically, suppose that $N$ agents all possess the same unprotected attribute(s) (e.g., they are bank customers with the same wealth classification). Then, the probabilistic control law (e.g.\ regression) governing the dependence of the agents' actions on the AI system output should be independent of the protected attribute(s) of the $ N$ agents (e.g.\ their ethnic group).
\end{itemize}

\begin{figure}[t!]
\centering
\includegraphics[width=0.85\columnwidth]{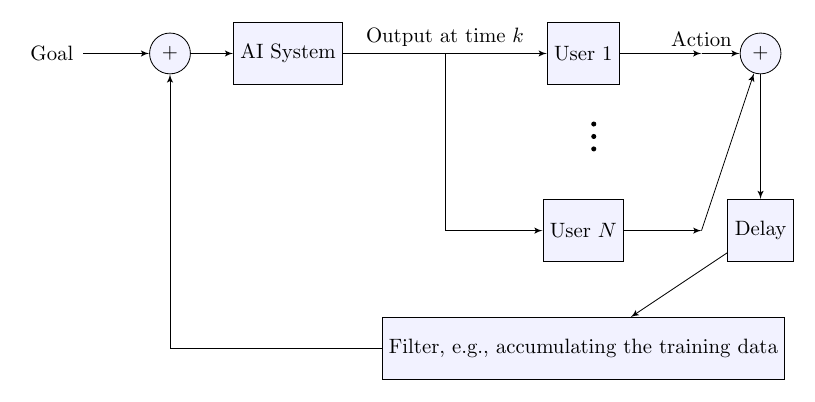}
\caption{A closed-loop model of an AI system and its interactions with agents: the AI system provides some outputs, e.g.\ scorecards in credit scoring, matches in a matching market, or suggestions in a decision support system. The agents observe the AI system output and take action in response. With some delay, these agent actions  are utilized in retraining the AI system.}
\label{fig:closed-loop1}
\end{figure}

\begin{figure}[t!]
\centering
\includegraphics[width=0.85\columnwidth]{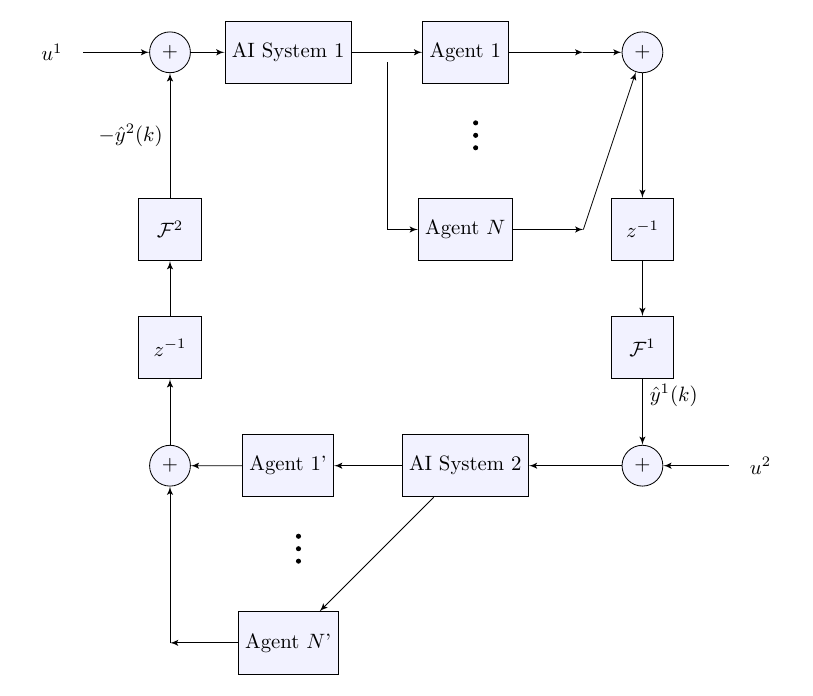}
\caption{A closed-loop model of an interconnection of two AI systems in a cascade, which---for example---can model a two-sided market in the sharing economy. In this case,  AI System 1 incentivises one side of the market (e.g.\ drivers), while AI System 2 incentivises the other side of the market (e.g.\ ride-seeking customers).}
\label{fig:closed-loop2}
\end{figure}

Ultimately, the repeated uses of an AI system can be modelled as a closed-loop (Figure \ref{fig:closed-loop1}).
The AI system produces an output with discrete time index, $k\in {\mathbb N}$, e.g.\ lending decisions in financial services, matches in a two-sided market, or suggestions in a decision-support system. 
The output is taken up by $N$ agents of the system (see the last item above), described respectively by their internal states, $x_i (k)$, $i\in[N]$ (here, $[N] \equiv \{1,\ldots,N\}$).
The agents respectively take some action, $y_i (k)$, which can be modelled via a probabilistic regression onto the AI system output and the agent's state, $x_i (k)$. 
In the sequel, we will assume the random variables (rvs), $y_i\left(k\right)$, are scalars, but this can be easily generalized. 
The aggregate of the agents' actions,  $y(k)=\sum\limits_{i = 1}^{N} y_i\left(k\right)$, at time $k$ is therefore also a rv. 
The AI system may not have access to either $x_i\left(k\right)$ or $y_i\left(k\right)$, but only to the processed (i.e.\ filtered) quantity, $y(k)$. 
The filter may accumulate the data, for instance, before filtering out anomalies. 

Following~\cite{10555099}, we now consider two alternative definitions of fairness in the composite agents-AI system (Figure~\ref{fig:closed-loop1}). Either definition is necessary for fairness in the conditional independence sense, since---if the conditional independence definition (final item above) is true---then either of these definitions is true. They are consistent with the fairness definitions studied in~\cite{gordaliza2019obtaining}, but are applied, here,  to {\em each\/} unprotected-attribute-indexed class, to accommodate the more nuanced fairness notion of conditional independence. 


The first definition evokes equal treatment in a single interaction with the AI system:

\begin{definition}[Equal treatment for each unprotected-attribute-indexed class of agents, $C$ ]
For each of the agents with common values of unprotected attributes, forming a class $C$, we require that 
the AI system induce 
the same expected output, $\pi(k)$, for each agent, $i\in C$, and that there exist a constant, $\overline{q}$, such that
\begin{equation}\label{eq:ch3-treatment}
\mathbb{E}[y_i(k)]  = \overline{q}(k),\;\;i\in C,
\end{equation}
and $\overline{q}(k)$ is independent of the initial conditions. 
\label{def:treat}
\end{definition}

One could obviously extend this to a variety of probabilistically approximately correct (PAC) versions of equal treatment. 
In contrast, the notion of equal impact involves the repeated interactions with the AI system and the long-run behaviour thereof: 

\begin{definition}[Equal impact for each unprotected-attribute-indexed class of agents, $C$]
For each of the agents with common values of unprotected attributes, $i\in C$, we require that 
there exist a constant, $\overline{r}_i$, such that 
\begin{equation}\label{eq:long-term-average}
\lim_{k\to \infty} \frac{1}{k+1} \sum_{j=0}^k y_i(j) = \overline{r}_i =\overline{r},
\end{equation}
and these limits are independent of the initial conditions.
\label{def:impact}
\end{definition}

Again, one could consider PAC variants. 
A necessary condition for (\ref{eq:long-term-average}) to hold is robustness of repeated interactions with the interconnection, in the following, ergodic sense:

\begin{definition}[Robustness in repeated interactions, in the ergodic sense, for each unprotected-attribute-indexed class of agents, $C$]
For all $N$ agents, $i\in C$,  we require that there exist a unique measure of the random variable, $y_i(k)$, that the closed-loop converges to, weakly  \cite{bogachev2018weak}, over repeated interactions.  
\label{def:robust}
\end{definition}

We will explain the condition in Definition \ref{def:robust} in more detail in Section~\ref{sec:ergo}. 

\section{Related Work}


\subsection{Toolkits}

Several PyTorch-based toolkits have been proposed recently. Our {\tt interconnect} toolkit focusses on the stochastic control aspects of a multi-agent AI system (Figure~\ref{fig:closed-loop1}) and guarantees equal impact for an interconnection (Figure~\ref{fig:closed-loop2}). As already stated, this is required by both  human rights legislation and by recent  AI regulation, but has not been addressed in any of the previous toolkits. 

Chopra et al.\  \cite{chopraagenttorch,chopra2024flame} proposed  both {\em AgentTorch\/} and {\em Flame}. The former focusses on the simulation of related systems, but without any grounding in ergodic properties.
Meanwhile, {\em Flame\/} focusses on the optimization aspects of the setting. 
Both our toolkit and these toolkits of Chopra et al.\ address scalability, with the capability to execute on clusters featuring both CPUs
and GPUs, and utilizing the underlying primitives of PyTorch. This makes it possible to scale to large-scale populations.

Our contribution is related to work on governance for multi-agent systems, both recent  \cite{shavit2023practices} and earlier. 
One can also view our work also as an example of automating the design of an agentic system \cite{hu2024automated}, and can be used in calibrating the multi-agent system \cite{quera2023bayesian}. A key recent development in this area is OpenAI's {\em Swarm\/} (\url{https://github.com/openai/swarm}).

\subsection{Ergodicity as a Prerequisite for Fairness}
\label{sec:ergo}

Even though our work builds on a long history of analysis of stochastic systems, we exploit results and methods from an area of stochastic processes that is relatively unknown in many mainstream areas of engineering analysis, i.e.\ the theory of {\em iterated function systems} \cite{Ramen2019,Elton1987,Barnsley1989,Diaconis1999}.
These constitute  a class of stochastic dynamic systems that became popular in the context of fractal image compression \cite{Elton1987} in the 1980s, and which goes further back,  to work on random ordinary differential equations \cite{strand1970random} from  1970. Roughly speaking, they describe a class of systems, in which state-to-state maps are drawn probabilistically, subject to specific distributions. These distributions may depend on the state of the system, in which case the probabilities are said to be state-dependent.\newline 

Recently, iterated function systems have arisen in systems where populations of agents---such as the agents interacting with the AI system in Figure~\ref{fig:closed-loop1}----evolve probabilistically, such as in the study of internet congestion protocols and sharing-economy systems, where these probability distributions are used to model the behaviour of rational or even non-rational agents. In this latter context, there may be a connection with classical {\em prospect theory}. 

Our interest in iterated function systems stems from a number of sources. First, they are an extremely powerful method of finite-scale modelling of population dynamics, providing a bridge between microscopic modelling and classical fluid limits. Second, powerful analytic results exist that can be exploited to deduce properties of populations of agents. An example arises when  deducing the ergodic behaviour of a given system, i.e.\  convergence of the stochastic process in some sense, independent of the initial conditions.  Ergodicity is important in many settings, for instance when deducing properties of a stochastic system based on  simulations. 
This property is also important when writing contracts in an economic setting. In our setting, a basic ergodicity property is a prerequisite to deducing fairness in the context of constrained resource allocation.  In the context of iterated function systems, the ergodic property reduces to basic properties on the probability functions, such as Lipschitz continuity, and boundedness away from 0 and 1. Finally,  there is a strong connection  between iterated function system dynamics and the behaviour of ensembles of agents with particular probabilistic behaviours.\newline 

Returning to our setting (Section~\ref{sec:setup}), the composite closed-loop involving agents and an AI system (Figure~\ref{fig:closed-loop1}) is an example of an iterated function system. The  assumptions of contraction-on-average are then sufficient to establish ergodicity, guaranteeing a form of robustness in  repeated interactions.
The fairness criteria posited by Definitions~\ref{def:treat} and \ref{def:impact} require this ergodicity to hold, in order for the long-run averages to exist.
A particular difficulty in our setting is, however, that the continuity assumptions---which would make it possible to reason about the contractivity-on-average---are often not satisfied.  
For example, classification tasks typically involve discrete decision/action/output sets, such as ``credit denied'' or ``credit approved'', involving finite state-space agents.  Assume, therefore, a common---i.e.\ $i\in[N]$-invariant---finite-cardinality, finite-dimensional  state space, $\mathbb{X}$, for each of the  $N$ agents (though this assumption can be relaxed): 
\begin{equation}\label{eq:action}
x_i(k) \in \mathbb{X}_i \equiv \mathbb{X} \equiv \{s_1,\ldots,s_L \}\subset \mathbb R^{n}.   
\end{equation}
Each agent's  state dynamics within (\ref{eq:action}) are subject to  feedback from the AI system. Specifically, at every discrete time, $k\in {\mathbb N}$, each agent receives the same feedback signal from the
AI system, $\pi_i(k) \equiv \pi(k) \in \Pi$ (Definition~\ref{def:treat}). The state transition dynamics of each of the $i\in[N]$ agents are assumed to be of $1\leq \tau_i < \infty$ types, each defined via a state transition map (subject to appropriate assumptions):
\begin{equation}
w_{ij}: \mathbb R^{n} \to \mathbb R^{n}, j=1,\ldots,\tau_i
\label{eq:statemap}
\end{equation}
We model the effect of the feedback signal, $\pi(k)$, on each of the $N$ agents stochastically, statically and discretely. Specifically, let $L_i(k) \in \{1, \ldots, \tau_i\}$ be a discrete stochastic pointer, which is modelled conditionally, given $\pi(k)$:
\begin{equation}
    p_{ij}(k)\equiv \Pr[L_i(k) =j | \pi(k)],\; j=1,\ldots,\tau_i\label{eq:prob-1}
 \end{equation}
The evolution of each agent, $i\in[N]$, is modelled via the pointer, $L_i(k)=j$; i.e.\ given $\pi(k)$, then $L_i(k)$ points w.p.\ $p_{ij}(k)$ to the $j$th one of the $\tau_i$ state transition maps, $w_{ij}$ (\ref{eq:statemap}). Finally, given $w_{ij}$, then
\begin{equation}
 x_i\left(k+1 | j \right) \equiv  w_{ij}\left\{x_i(k)\right\}, \; j = 1, \ldots, \tau_i,\label{eq:state} 
\end{equation}
and we denote $x_i\left(k+1 | j \right) \to x_i\left(k+1  \right) $ in the next time step.

We model the state-to-output maps for each of the $i\in[N]$ agents in the same discrete, hierarchical manner; i.e.\  the output map is chosen from a finite set, with the choice again controlled by the AI system output (i.e.\  feedback signal), $\pi(k) \in \Pi$. 
 The finite-state output (i.e.\ action) space (Section~\ref{sec:setup}) of each of the $i\in[N]$ agents is once again assumed to be common to the $N$ agents (though, again, this assumption can be relaxed):
\begin{equation}\label{eq:demands}
y_i \in \mathbb{Y}_i \equiv {\mathbb Y} \equiv \{ a_1, \ldots, a_m\} \subset {\mathbb R}^q.
\end{equation}
Once again, the controlled dynamics---here, between $x_i$ and $y_i$---are modelled hierarchically via a discrete pointer, $M_i(k)$, to $1\leq \kappa_i < \infty$ output maps: 
\begin{equation}
w'_{i\ell}: \mathbb R^{n} \to{\mathbb R}^q,\; \ell= 1,\ldots,\kappa_i.
\label{eq:opmap}
\end{equation}
The pointer, $M_i(k)$, to the output map of the $i$th agent at discrete time, $k$, is once again realized conditionally on the AI system output (i.e.\ feedback signal), $\pi(k)$:
\begin{equation}
    p'_{il}(k)\equiv \Pr[M_i(k) =l | \pi(k)], \;l=1,\ldots,\kappa_i\label{eq:prob-2}
 \end{equation}
The controlled output, $y_i(k)$, of each agent, $i\in[N]$, is again modelled hierachically, this time  via the pointer, $M_i(k)=l$; i.e.\ given $\pi(k)$, then $M_i(k)$ points w.p.\ $p'_{il}(k)$ to the $l$th one of the $\kappa_i$ output  maps, $w'_{il}$ (\ref{eq:opmap}). Finally, given $w'_{il}$, then
\begin{equation}
 y_i\left(k | l \right) \equiv  w'_{il}\left\{x_i(k)\right\},  l = 1, \ldots, \kappa_i,\label{eq:output} 
\end{equation}
and we denote $y_i\left(k | l \right) \to y_i\left(k  \right) $ in any subsequent multi-agent filtering step (\ref{eq:long-term-average}).




This discrete, hierarchical modelling of the closed-loop interactions between the agents and the AI system (Figure~\ref{fig:closed-loop1}) ensures that the graph, 
$G = (X, E)$, is strongly connected. Then, vitally, there
exists an invariant measure for the feedback loop. If, in
addition, the adjacency matrix of the graph is primitive,
then the invariant measure is attractive and the system
is uniquely ergodic.

For linear systems, this is a direct consequence of \cite{Werner2004,werner2005contractive} and the fact that the necessary contractivity properties follow from the internal asymptotic stability of controller and filter.
For non-linear systems, similar results can be obtained using \cite[Theorem 2]{Marecek2021}. 

\section{Contraction on Average}


Let us consider a Markov chain on a state space representing all the system components. 
In particular, let us consider a model of the state space, $\mathbb{X}$, which combines the states of the agents across all populations, controllers, and filters.  
Then, one can model the evolution of the system with a stochastic difference inclusion (SDI):
\begin{equation}
\label{eq:generalSDI}
x(k+1) \in \{ F_m(x(k)), \, m \in {\mathbb M}\}.
\end{equation}
Here, the state, \(x(k)\), consists of signals at various points in the system, and the \(F_m\) denote the transitions determined by the agents' stochastic logic and the controller-filter dynamics. 
The indices, $m\in\mathbb{M}$, of the maps, $F_m: \mathbb{X} \to \mathbb{X}$, are   
sampled via a specified probability mass function, $p_m$. 
In the  general case, note that the ordering of the
states of the agents across all populations, controllers, and filters can be
obtained with a breadth-first search (BFS) on the graph of the system,
which then defines $F_m$.

Finally, we say that the Markov chain---or, equivalently, the stochastic difference inclusion---is \emph{ergodic} (i.e.\ there exists a unique invariant measure) if \eqref{eq:generalSDI}  satisfies the contraction-on-average condition, i.e.\ if there there exists a constant, 
$0<C<1$, such that
\begin{equation}
\label{eq:average-contraction1}
\sum_{m\in{\mathbb M}}p_m \frac{\|F_{m}(x)-F_{m}(\hat{x})\|}{\|x-\hat{x}\|} < C \quad \forall x, \;
\hat{x} \in \mathbb{X}.
\end{equation}

 In the {\tt interconnect} toolkit, the  {\tt get\_factor\_from\_list()} function (Figure~\ref{fig:contraction-factor1a})  provides a stochastic approximation of the 
the contraction-on-average factor, $C$, in (\ref{eq:average-contraction1}). This function takes the set of admissible reference signal values
and the set of probabilities in the agents (realizations of uncertainty), and runs simulations with 
the given parameters. 


\begin{figure}[tbh]
\begin{lstlisting}[language=Python, basicstyle=\small]
C = get_factor_from_list(
    reference_signals=reference_signals,
    agent_probs=np.array([[eps, 1.0-eps], [eps, 1.0-eps]]), sim_class=ExampleReLUSim,
    it=200, trials=500,
    node_outputs_plot="A1",
    show_distributions_plot=True,
    show_distributions_histograms_plot=False)
\end{lstlisting}
\caption{An example of computing the contraction-on-average factor, $C$, for the system
 model in Figure~\ref{fig:closed-loop1c} (Supplementary Material).}
\label{fig:contraction-factor1a}
\end{figure}


\begin{figure}[tbh]
\includegraphics[width=0.85\columnwidth]{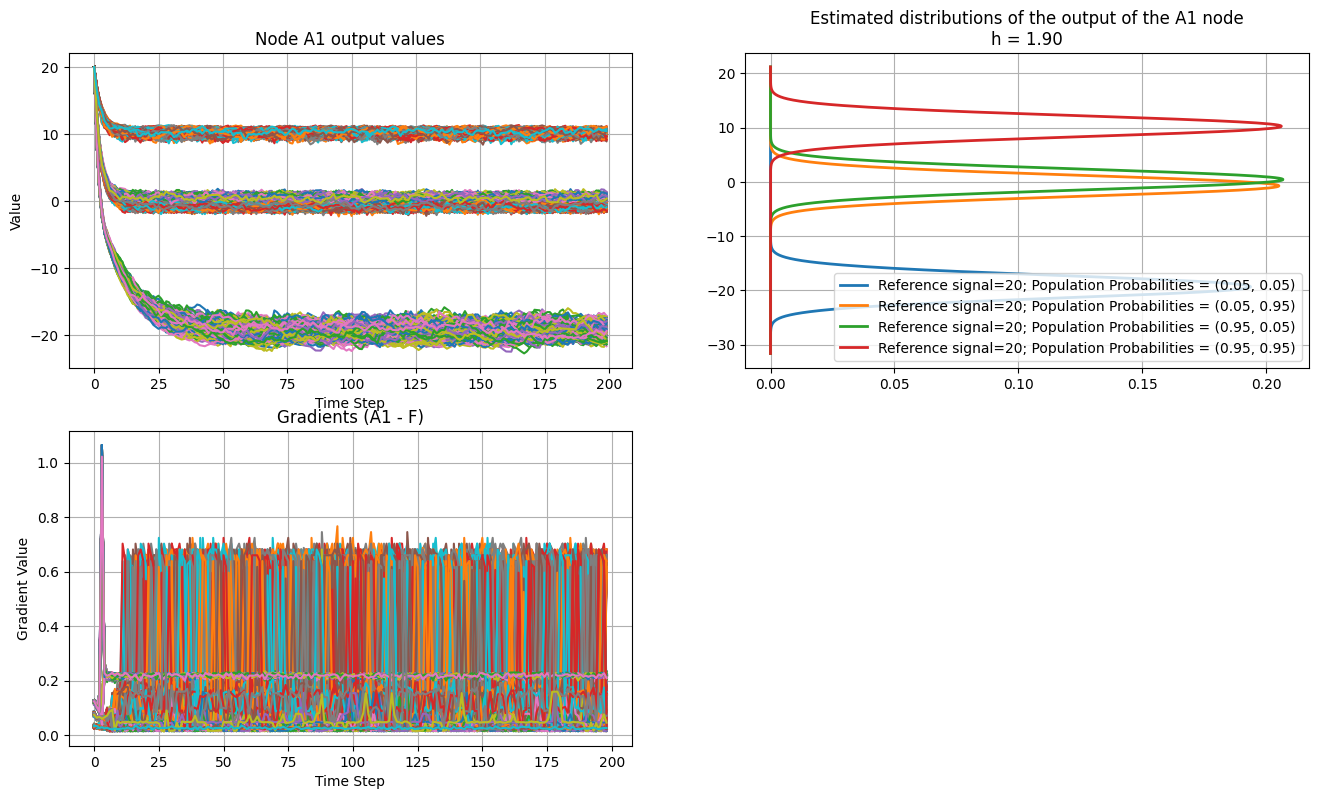}
\captionof{figure}{Plots generated during testing with different combinations of {\tt reference\_signals} and agent inner probabilities for the system model in Figure~\ref{fig:closed-loop1c}.
}
\label{fig:conraction-factor1b}
\end{figure}

\paragraph{Closed-loop System Example}
Consider, for example, the state, \(x(k)\), and the transitions, \(F_m\), specified as follows (\ref{eq:generalSDI}):

\begin{enumerate}
    \item The input signal, \(r(k)\) (reference), is passed to the error computation node (\(A_1\)) to compute the error, \(e(k) = r(k) - f(k)\), where \(f(k)\) is the output of the filter.
    \item The controller processes \(e(k)\) according to the PI logic,
    \begin{equation}
    \pi(k) = K_p e(k) + K_i \sum_{\ell=0}^k e(\ell),
    \end{equation}
    producing the control output, \(\pi(k)\).
    \item The control signal, \(\pi(k)\), is sent to populations \(P_1\) and \(P_2\). Each population, $p\in\{1,2\}$, consists of agents with stochastic behaviour,
    \begin{equation}
    y_p(k) = \sum_{i=1}^{N_p} h_{j(i)}(\pi(k)),
    \end{equation}
    where $h_{j(i)}(k)$ is the output of the agent using one of its functions, $j$, randomly selected via a pre-specified probability function.
    \item The aggregated agent output, \(y(k)\), is then  filtered via $\mathcal{F}$:
    \begin{equation}
    f(k) = \mathcal{F}(y(k)).
    \end{equation}
\end{enumerate}


\begin{figure}[tbh]
\centering
\includegraphics[width=0.65\columnwidth]{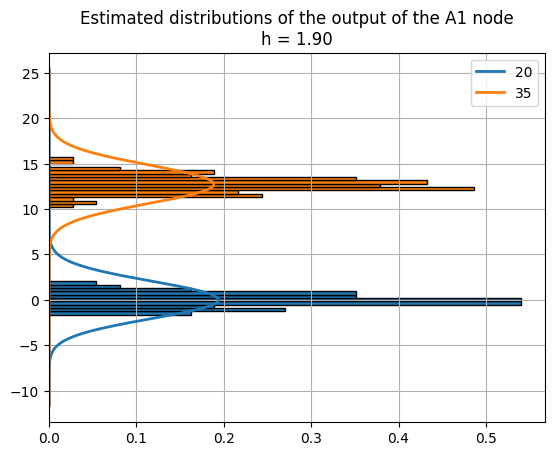}
\captionof{figure}{Plots of the estimated distributions of the model state after 100 iterations, for the model in Figure~\ref{fig:closed-loop1c}.
}
\label{fig:distibution-estimation}
\end{figure}

Finally, the  evolution of the system can be represented as
\begin{equation}
\label{eq:updated-F_m-definition}
F_m(x(k)) := \left[\begin{array}{rll}
r(k) &&\\
e(k) &=& r(k) - f(k) \\
\pi(k) &=& K_p e(k) + K_i \sum_{\ell=0}^k e(\ell) \\
y(k) &=& \sum_{i=1}^{N_{P1}} h_{j(i)}(\pi(k)) + \sum_{i=1}^{N_{P2}} h'_{j(i)}(\pi(k)) \\
f(k) &=& \mathcal{F}(y(k))
\end{array}\right].
\end{equation}
The maps, $F_m$ (\ref{eq:generalSDI}), are indexed from the set,
\begin{equation}
\mathbb{M} := \{ j_1, \ldots j_{N_{1}} \} \times \{j'_1, \ldots j'_{N_{2}} \}.
\end{equation}
The probability of choosing map, $F_m$, is:
\begin{equation}
\mathbb{P}(x(k+1) = F_m(x(k))) = 
\prod_{i=1}^{N_{1}} p_{j_i} \prod_{i=1}^{N_{2}} p'_{j'_i}.
\end{equation}
where $p_{j_i}$ and $p'_{j_i}$ are the probabilities of choosing function $j$ for agent $i$ in the respective populations, $P_1$ and $P_2$ (step~3 above).


The {\tt interconnect} toolkit can be used both to check the contraction on average condition is satisfied (cf. Figure~\ref{fig:contraction-factor1a}), and to approximate the   probability distribution of the output signals in the checkpoint node using kernel density estimation (KDE):
\begin{equation}
    \hat{f}_h(z) = \frac{1}{n} \sum_{i=1}^{n} K_h(z - z_i) = \frac{1}{n h} \sum_{i=1}^{n} K\left(\frac{z - z_i}{h}\right).
\end{equation}
Here, $K$ is a symmetric elementary distribution (i.e.\ kernel),  $h$ is the kernel size (manually set in the toolkit), 
$\{z_1, ..., z_n\}$ is a set of independent and identically distributed measurements, and $z$ is the point at which we want to estimate the density.
The KDE function accepts a set of acceptable starting node values and estimates a distribution for each of them based on the specified number of simulation runs (Figure~\ref{fig:distibution-estimation}).

\section{Conclusions and Limitations}

We have presented the open-source toolkit, {\tt interconnect}, which uses techniques from stochastic control in the study of multi-agent systems. Specifically, this toolkit makes it possible to study the robustness of closed loops and related interconnections of AI systems, and related notions of fairness.

The toolkit's key limitation is related to the scalability of the estimation of constant $C$ in the contraction-on-average test \ref{eq:average-contraction1}. There are methods with bounded error that scale to well-known test-cases such as MNIST for depths and widths of the neural networks in the hundreds \cite{wang2024scalability}, and methods that scale to any instances, with any depths and any widths \cite{weng2018evaluating,goodfellow2018gradientmaskingcausesclever,weng2018extensions}, but whose error has not been bounded \emph{a priori}. 
We imagine that \cite{weng2018evaluating,goodfellow2018gradientmaskingcausesclever,weng2018extensions} could benefit from importance sampling and related variance-reduction techniques.  
Future work on the challenge of scalability would be most welcome.


\FloatBarrier
\bibliographystyle{ieeetr}
\bibliography{main}

\newpage

\clearpage
\section*{Supplementary Material: Part A}
\subsection*{A P\MakeLowercase{y}T\MakeLowercase{orch} Implementation}

It is clear that a formal specification of the AI system and the environment \cite{seshia2018formal} and their interactions is needed to verify the properties of the closed loop. Considering the complexity of many AI systems, it seems natural to use the specification of the AI system within  a machine-learning library such as PyTorch and their computational graphs.

We have implemented a toolkit on top of PyTorch, which makes it possible to specify all elements of the closed loop, such as in Figures 
\ref{fig:closed-loop1} and \ref{fig:closed-loop2}, as elements of the computational graph, and then utilize automatic differentiation within PyTorch ({\texttt torch.autograd}). Each element is specified using an instance of a class such as in Figure \ref{fig:class}.
For example, when one wishes to specify a simple filter, one may consider the snippet of Figure \ref{fig:filter}.

\begin{figure}[ht!]
\begin{lstlisting}[language=Python, basicstyle=\tiny]
class YourLogicClass:
 def __init__(self, param1, param2):
  self.variables = ["input1", "input2", ...]
  self.tensors = {"intpit1": torch.tensor([0.0], requires_grad=True), ...}
  # Initialize other necessary attributes

 def forward(self, values):
  output = None 
  # Compute the output based on input values
  # Update internal state if necessary
  return output
\end{lstlisting}
\caption{Specifying one element of the closed loop.}
\label{fig:class}
\end{figure}

\begin{figure}[ht!]
\begin{lstlisting}[language=Python, basicstyle=\tiny]
class FiltererLogic:
 def __init__(self):
  self.tensors = {"S": torch.tensor([0.0], requires_grad=True),
   "K": torch.tensor([2.0], requires_grad=True)}
  self.variables = ["S"]
    
 def forward(self, values):
   self.tensors["S"] = values["S"]
   result = - (self.tensors["S"]) / self.tensors["K"]
   return result
\end{lstlisting}
\caption{Specifying a filter in our interconnect.}
\label{fig:filter}
\end{figure}

Notice that the forward method and PyTorch's automatic differentiation engine ({\texttt torch.autograd}) can be utilized in a number of ways:
\begin{itemize}
\item bounding the moduli of (Lipschitz, Dini) continuity of individual elements of the interconnection, if deterministic, or expressing the worst-case modulus of continuity with respect to the realization of uncertainty. If all of the elements are contractive in this setting, we can guarantee contractivity of the whole loop, and thus (trivially) the existence of the limit \eqref{eq:long-term-average} used in Definition \ref{def:impact} of equal impact. 
\item bounding the expectation of the modulus of (Lipschitz, Dini) continuity of the closed loop or more complicated interconnection, with respect to the uncertainty involved. If we can guarantee this contractivity-on-average for the whole loop, non-trivially, we can guarantee the existence of the limit \eqref{eq:long-term-average} used in Definition \ref{def:impact} of equal impact. The details have been developed \cite{Fioravanti2019,Ramen2020,ghosh2021ergodic,Marecek2021,Slava2021} in a number of settings, recently.
\item optimization of parameters of the elements in the interconnection, either with respect to some extrinsic objectives (e.g., financial), or with respect to the ergodic properties (e.g., mixing rate \cite{mathew2011metrics}, which is related to the  contractivity-on-average above, cf. \cite{ghosh2022iterated}). Perhaps the most useful element within the interconnection to train with such objectives is, obviously, the AI system itself, but one may also wish to calibrate \cite{quera2023bayesian} the model of population, for instance. 
\item simulation of the  interconnection. While one can clearly benefit from approximating the gradients within simulaton \cite{chopraagenttorch}, simulation is most useful in cases, where the ergodic properties (cf. \cite{ghosh2022iterated}) are satisfied. Indeed, the existence of large limits, such as \eqref{eq:long-term-average} makes the simulation into a consistent estimator of such long-run properties.
In contrast, when the ergodic properties are not satisfied, the consistency of any estimator is a non-trivial question.
\end{itemize}

\subsection{Another Illustrative Example}

\begin{figure}[t!]
\includegraphics[width=0.85\columnwidth]{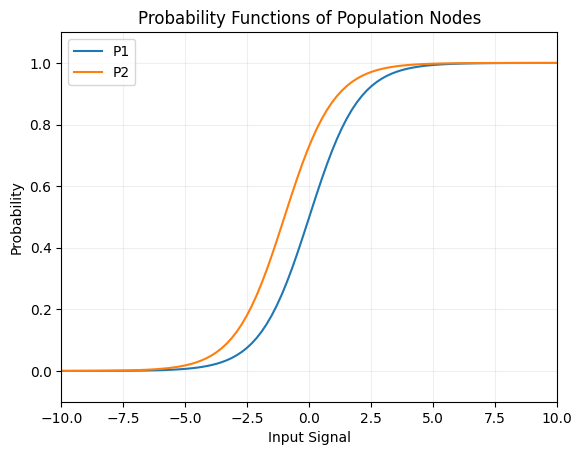}
\captionof{figure}{Response functions of the two types of agents.}
\label{fig:agent-functions}
\end{figure}

Let us now consider a complete, and slightly  more elaborate example, where we the AI system uses REctified Linear Unit (ReLU) in a deep neural network (DNN), and there are two types of agents, whose probabilistic response functions are captured in Figure \ref{fig:agent-functions}.

One of the aims of the toolkit is to make the specification as easy as possible. With commonly considered filters and controllers pre-specified, the non-trivial example of Figure \ref{fig:agent-functions} can be captured in several dozen lines, featuring the snippet in Figure~\ref{fig:closed-loop1b}.
Notice that one can mix in-built classes such as Delay, as well as custom ones, such as FilterLogic of Figure~\ref{fig:filter}.

\begin{figure}[tbh]
\begin{lstlisting}[language=Python, basicstyle=\small]
refsig = ReferenceSignal(name="r")
refsig.set_reference_signal(torch.tensor(reference_signal, requires_grad=True))
agg1 = Aggregator(name="A1", logic=self.AggregatorLogic1())  # Error computation
agg2 = Aggregator(name="A2", logic=self.AggregatorLogic2())  # Sums population outputs
cont = Controller(name="C", logic=ReLUControllerLogic())
pop1 = Population(name="P1", logic=AgentLogic(), number_of_agents=20)
pop2 = Population(name="P2", logic=AgentLogic(offset=0.4), number_of_agents=20)
delay = Delay(name="Z", logic=ESDelayLogic())
fil = Filter(name="F", logic=self.FilterLogic())
self.system.add_nodes([refsig, agg1, agg2, cont, pop1, pop2, delay, fil])
self.system.connect_nodes(refsig, agg1)
self.system.connect_nodes(agg1, cont)
self.system.connect_nodes(cont, pop1)
self.system.connect_nodes(cont, pop2)
self.system.connect_nodes(pop1, agg2)
self.system.connect_nodes(pop2, agg2)
self.system.connect_nodes(agg2, delay)
self.system.connect_nodes(delay, fil)
self.system.connect_nodes(fil, agg1)

self.system.set_start_node(refsig)
self.system.set_checkpoint_node(agg1)
\end{lstlisting}
\caption{An example specification of a closed loop corresponding to model of Figure~\ref{fig:closed-loop1}.}
\label{fig:closed-loop1b}
\end{figure}

\begin{figure}[tbh]
\includegraphics[width=0.45\columnwidth]{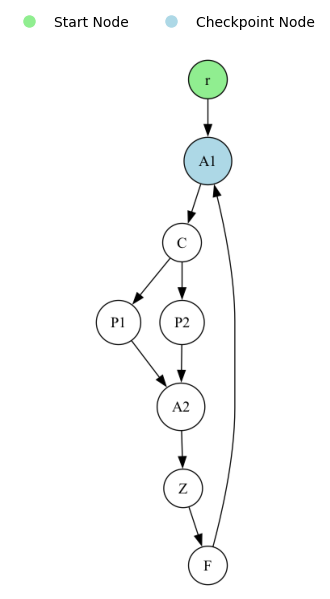}
\captionof{figure}{A visual representation of the interconnection captured in Figure~\ref{fig:closed-loop1b} generated by the toolkit.
}
\label{fig:closed-loop1c}
\end{figure}


As another example, one could consider the model of Figure~\ref{fig:closed-loop2}, which can be implemented in only a handful of extra lines. Its rendering from the {\tt interconnect} toolkit is provided in Figure~\ref{fig:system-graph-f2}, and the notebook is attached in the supplementary material. 

\begin{figure}[tbh]
\includegraphics[width=0.65\columnwidth]{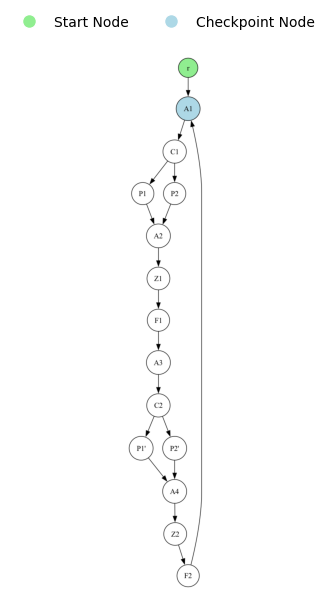}
\caption{A visual representation of the interconnection captured in Figure~\ref{fig:closed-loop2} generated by the toolkit.}
\label{fig:system-graph-f2}
\end{figure}


\end{document}